\pdfoutput=1

\documentclass[11pt]{article}

\usepackage[final]{latex/acl}

\usepackage{times}
\usepackage{latexsym}
\usepackage{subfig}

\usepackage[T1]{fontenc}

\usepackage[utf8]{inputenc}

\usepackage{microtype}

\usepackage{inconsolata}

\usepackage{graphicx}
\usepackage{booktabs} 
\usepackage{multirow}

\title{GPT vs RETRO: Exploring the Intersection of Retrieval and Parameter-Efficient Fine-Tuning}

\author{Aleksander Ficek\thanks{Equal contribution.}, Jiaqi Zeng\footnotemark[1], Oleksii Kuchaiev \\
NVIDIA\\
\texttt{\{aficek,jiaqiz,okuchaiev\}@nvidia.com} \\
}

\usepackage{booktabs}
\usepackage{stackengine}

\begin{document}
\maketitle
\begin{abstract}

Parameter-Efficient Fine-Tuning (PEFT) and Retrieval-Augmented Generation (RAG) have become popular methods for adapting large language models while minimizing compute requirements. In this paper, we apply PEFT methods (P-tuning, Adapters, and LoRA) to a modified Retrieval-Enhanced Transformer (RETRO) and a baseline GPT model across several sizes, ranging from 823 million to 48 billion parameters. We show that RETRO models outperform GPT models in zero-shot settings due to their unique pre-training process but GPT models have higher performance potential with PEFT. Additionally, our study indicates that 8B parameter models strike an optimal balance between cost and performance and P-tuning lags behind other PEFT techniques. We further provide a comparative analysis between applying PEFT to an Instruction-tuned RETRO model and base RETRO model. This work presents the first comprehensive comparison of various PEFT methods integrated with RAG, applied to both GPT and RETRO models, highlighting their relative performance.

\end{abstract}

\section{Introduction}

Pre-trained large language models have made a demonstrable impact across applications in academia and industry. Many use cases, however, require LLMs adapted to specific tasks and unique information but lack the resources for extensive retraining. To address this, Parameter-Efficient Fine-Tuning (PEFT) \citep{peft_survey} and Retrieval-Augmented Generation (RAG) \citep{rag_survey} have become popular methods due to their effectiveness and efficiency, inspiring new lines of research.

\begin{figure}[ht]
    \hspace*{-0.35cm}
    \vspace{-0.49cm}
    \includegraphics[scale=0.66]{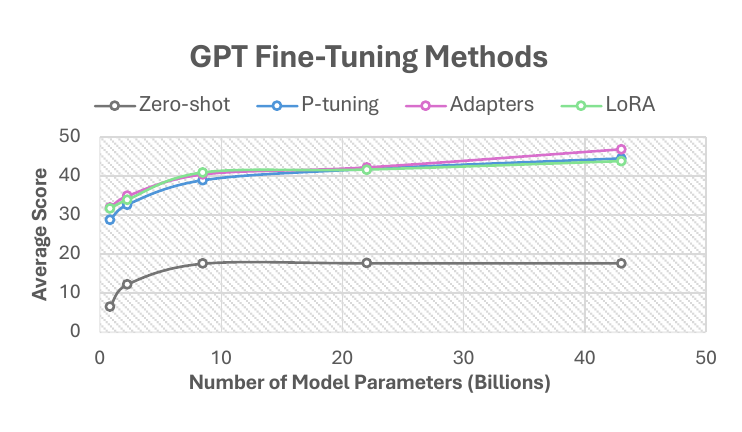}
    \hspace*{-0.35cm}
    \includegraphics[scale=0.66]{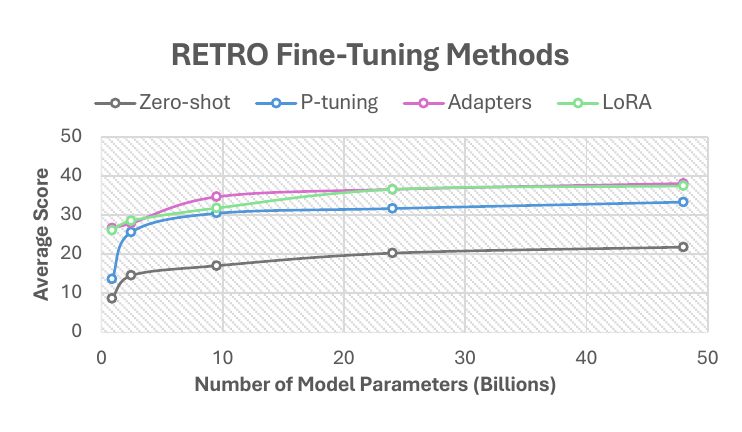}
    \vspace{-2mm}
    \vspace{-5mm}
    \caption{Average GPT vs RETRO scores of six datasets across model sizes of 823M to 48B parameters.}
    \label{fig:peft_methods}
    \vspace{-5mm}
\end{figure}

\vspace{1mm}
PEFT has been proven to be a comparable substitute to Supervised Fine-Tuning (SFT) by achieving competitive performance at a fraction of the number of updated parameters \citep{peft_survey}. In this paper we select P-tuning \citep{ptuning}, Adapter modules \citep{adapters} and Low-Rank Adaptation (LoRA) \citep{lora} as representative PEFT methods. P-tuning involves training continuous prompt embeddings to guide output for specific tasks without modifying base model parameters. Adapters operate by training fully connected layers inserted throughout the base model while keeping the remaining parameters frozen. LoRA further decomposes the inserted layers into low-rank matrices, enhancing efficiency.

\begin{figure*}[ht]
    \hspace*{-0.2cm}
    \centering
    \includegraphics[scale=0.61]{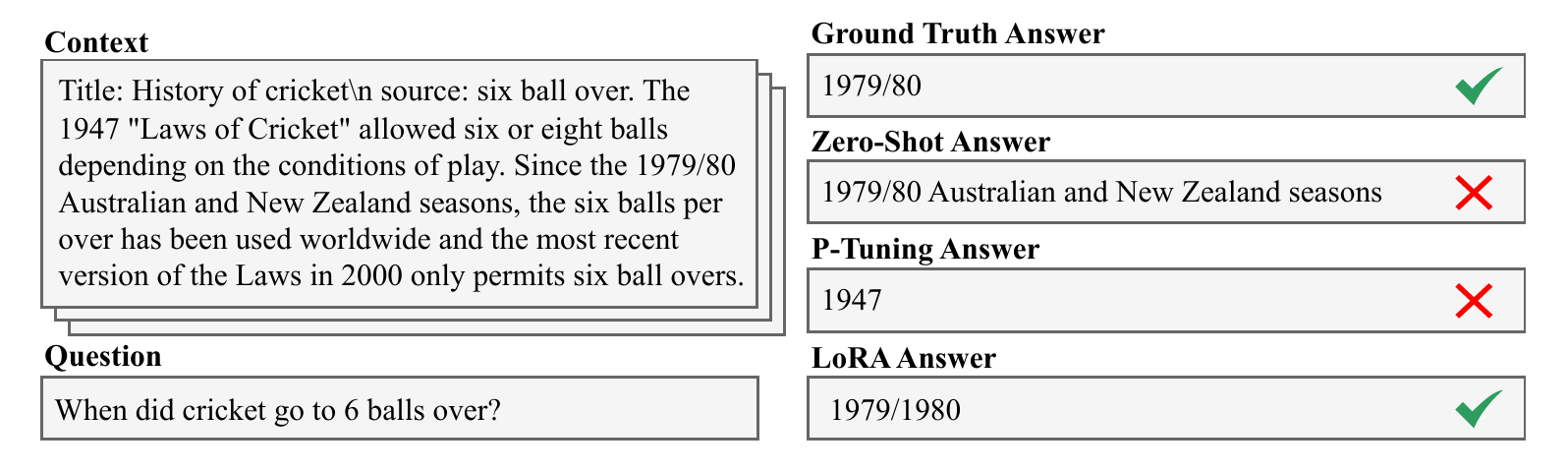}
    \vspace{-5mm}
    \caption{Sample entry inputs and outputs from NQ dataset}
    \label{fig:sample_entry}
    \vspace{-3mm}
\end{figure*}


Retrieval-augmented generation (RAG) improves model quality by incorporating external knowledge through mechanisms like BM-25 or TF-IDF \citep{tfidf}, online web search \citep{pagerank}, or trained dense retriever models \citep{retriever}. Any LLM can be transformed into a retrieval-augmented model by concatenating retrieved sources with the input query, provided it fits within the model's context window. \citet{nvidia_context_retrieval} found that retrieval significantly improves GPT model quality on long context tasks, reducing the "lost in the middle" effect \citep{lostinthemiddle} and offering inherent efficiency benefits.

Alternatively, there exist multiple works \citep{retro, realm, atlas, webgpt} that have integrated retrieval as part of model pretraining or fine-tuning to notable success when compared to typical GPT models despite being a much lesser explored domain. RETRO \citep{retro} is of particular interest due to its unique approach of incorporating a retrieval module directly into the transformer architecture via a chunked-cross attention mechanism and ability to scale to trillions of tokens resulting in reduced perplexity. Subsequently, \citet{nvidia_shall_we} showed that RETRO at sizes up to 9.5 billion parameters largely outperforms GPT on specific knowledge-intensive tasks. Furthermore, \citet{instructretro} illustrated that when scaled up to 48 billion parameters and instruction-tuned, RETRO performed better than equivalent GPT models on several question answering, reading comprehension and summarization tasks.

In this paper we continue the exploration of RETRO versus GPT through the lens of parameter efficient finetuning. We apply P-tuning, Adapter modules and LoRA to multiple tasks with retrieval for both RETRO and GPT models. To our knowledge, this paper provides the first in-depth comparison of various Parameter Efficient Fine-Tuning integrated with Retrieval-Augmented Generation, uniquely applied to both GPT and RETRO models.

\section{Related Work}

Previous works like \citet{reviewer1rebuttal}, have compared multiple PEFT methods but lacked comparison for retrieval-based tasks and retrieval augmented language models. In this section we focus on recent work that combine finetuning with retrieval. A comprehensive survey~\cite{rag_survey} synthetized multiple comparative studies on PEFT and RAG, underscoring the potential benefits of combining these approaches as a promising direction for future investigation. There are multiple works that provide methods to combine RAG with fine-tuning to improve accuracy \citep{finetuning_rag_1, finetuning_rag_2, finetuning_rag_3}. Multiple studies have explored the comparison between fine-tuning and retrieval. \citet{finetuning_vs_rag_1} and \citet{finetuning_vs_rag_2} reported improved accuracy using RAG over fine-tuning GPT models, while also noting suboptimal results when combining the two methods. \citet{finetuning_vs_rag_3} demonstrated improved outcomes by integrating both approaches for specific agriculture and geography tasks. Additionally, \citet{finetuning_vs_rag_4} compared the efficacy of these methods, including full and QLoRA fine-tuning \citep{qlora}, in low-frequency entity question-answering tasks. 
These studies collectively suggest the need for comprehensive investigation into multiple PEFT techniques combined with RAG and maintain retrieval pretrained LLMs with PEFT to be unexplored, thereby motivating our research.

\section{Experimental Setup}

\subsection{Datasets}

To cover several task categories, we use six datasets suited to benefit from retrieval and finetuning. We select \textbf{Natural Questions} (NQ) \citep{nq}, \textbf{TriviaQA} (TQA) \citep{triviaqa}, \textbf{NarrativeQA} (NQA) \citep{narrativeqa} and \textbf{Qasper} \citep{qasper} for document question answering, \textbf{QuALITY} \citep{quality} for multiple-choice question answering, and \textbf{QMSum} \citep{qmsum} for query-based summarization. Table \ref{tab:data-stat} details the sizes of dataset training, validation and test partitions. Each of these datasets contain necessary external knowledge that must be filtered via retrieval and response behaviour that encourages finetuning. Following the official metrics, we use F1 score for evaluating document QA, exact match for mutliple-choice QA and the geometric mean of ROUGE-1/2/L  \citep{rouge}  for summarization.

\begin{table}[ht]
\centering

\resizebox{\columnwidth}{!}{
\begin{tabular}{lccccccc}
\toprule
& NQ & TQA & NQA & QASPER & QUALITY & QMSUM \\
\midrule
Train & 79168 & 78785 & 44002 & 2053 & 2018 & 1005\\
Valid & 8757& 8837 &11001 & 514 & 505 & 252\\
Test &3610 & 11313 & 5859 & 1726 & 2086 & 272\\
\bottomrule
\end{tabular}
}
\caption{Number of samples in train/validation/test split for each dataset.}
\label{tab:data-stat}
\end{table}

\subsection{Models}

In order to understand the effect of model scaling, we use base GPT models of sizes 823M (Extra Small), 2.25B (Small), 8.5B (Medium), 22B (Large), and 43B (Extra Large), as introduced in \citet{instructretro}, which were pretrained on a massive dataset of 1.2 trillion tokens. We employ the corresponding RETRO models from the same work as the foundation for our retrieval pretrained LLM experiments. Notably, the RETRO architecture features an encoder that extracts neighbors from an external database, which increases the total model size to 877M, 2.47B, 9.5B, 24B, and 48B, respectively. 
\citet{instructretro} found ablating the encoder after pretraining led to comparable results. In our paper we include it so that adapter modules and LoRA layers are added throughout decoder and encoder components. We choose the GPT and RETRO model types for our experiments because they are representative architectures of the general and retrieval LLM landscape while allowing us to leverage the large pretrained models introduced in \citet{instructretro}. For more information on the base models we refer readers to the original work.

\subsection{Retrieval}
We follow~\citet{instructretro,nvidia_context_retrieval} to use Dragon+~\citep{lin2023train} as a retriever. Dragon+ is a dual encoder model that consists of a query encoder and a context encoder.
We first chunk each context document with 100 words, and then encode
both the questions and all chunks independently with corresponding encoders. The most relevant 5 chunks, ranked by the dot product of the question embedding and chunk embedding, are retrieved as neighbors. For GPT models, they are concatenated together (following the left to right order from the most relevant to least relevant) as the
context of the prompt for generation. For RETRO models, they interact with the question during generation through chunked cross-attention. We choose Dragon+ as the retriever because it was employed in the original RETRO paper \citep{retro} and has achieved decent performance in other works \citep{instructretro}. Here we are interested in relative performance between GPT and RETRO models, enabling comparison against the architectures instead of comparing multiple retrievers which we leave for future work.

\subsection{Parameter Efficient Fine-Tuning}

We implement P-tuning in RETRO akin to GPT. Virtual tokens are added to the beginning of the decoder. Based on the design of chunked cross-attention, left padding is added to ensure the length of input (virtual tokens + context + question) is a multiple of chunk size. Adapter and LoRA layers are in all attention layers in both transformer architectures. This means that for RETRO they are also inserted in the retrieval encoder which receives retrieved neighbors. We provide additional hyperparameter tuning, resource utilization and prompt template details in Appendix \ref{sec:appendixa}. We also include Table \ref{tab:num_params} for a full list of base and PEFT model parameter counts.

\begin{table}[ht]
\centering

\resizebox{\columnwidth}{!}{
\begin{tabular}{cccccc}
\toprule
 \textbf{Type} & \textbf{Size} & \textbf{Base Model} & \textbf{P-Tuning} & \textbf{Adapters} & \textbf{LoRA} \\
\midrule
\multirow{5}{*}{\textbf{GPT}} & Extra Small & 823M & 2.2M & 3.2M & 3.1M \\
                              & Small & 2.25B & 3.3M & 6.5M & 6.3M \\
                              & Medium & 8.5B & 5.6M & 18.8M & 16.8M \\
                              & Large & 22B & 8.0M & 35.2M & 31.2M \\
                              & Extra Large & 43B & 10.4M & 63.2M & 50.4M \\
\midrule
\multirow{5}{*}{\textbf{RETRO}} & Extra Small & 877M & 2.2M & 3.6M & 4.3M \\
                                & Small & 2.47B & 3.3M & 7.3M & 8.7M \\
                                & Medium & 9.5B & 5.6M & 20.8M & 22.4M \\
                                & Large & 24B & 8.0M & 43.5M & 42.4M \\
                                & Extra Large & 48B & 10.4M & 70.6M & 68.0M \\
\bottomrule
\end{tabular}
}
\caption{Base and PEFT model number of parameters}
\label{tab:num_params}
\end{table}

\begin{table*}[ht]
\centering
\renewcommand{\arraystretch}{1.2}
\Huge
\resizebox{\textwidth}{!}{
\begin{tabular}{clccccccccccccccccc}
\toprule
                            & & \multicolumn{2}{c}{\textbf{NQ}} &
                            \multicolumn{2}{c}{\textbf{TQA}}&
                            \multicolumn{2}{c}{\textbf{NQA}} &
                            \multicolumn{2}{c}{\textbf{QASPER}} & 
                            \multicolumn{2}{c}{\textbf{QUALITY}} &
                            \multicolumn{2}{c}{\textbf{QMSUM}}&
                            \multicolumn{2}{c}{\textbf{AVERAGE}} \\                           
                            & & GPT & RETRO & GPT & RETRO & GPT & RETRO & GPT & RETRO & GPT & RETRO & GPT & RETRO & GPT & RETRO         \\
\midrule
\multirow{4}{*}{\textbf{Extra Small}} & Zero-shot        & 2.95       & \textbf{8.28}   & 9.99   &\textbf{19.26} &\textbf{7.07} &4.87 &   9.00    &     \textbf{10.79}           & 0.38       & \textbf{0.48}    &\textbf{9.83}& 7.78& 6.54 &\textbf{8.58} \\ 
& P-tuning     & \textbf{24.74}       &7.60    &\textbf{63.63}   &24.61 &\textbf{16.74} & 6.69 &    \textbf{24.54}    &     11.94          & \textbf{24.59}      & 17.45    &\textbf{18.25} & 13.63 &\textbf{28.75} & 13.65  \\
 &Adapter        & \textbf{38.48}     &23.69   &\textbf{67.99}  &59.60 &\textbf{17.76} & 15.42 &    \textbf{23.52}    &     20.96        & 24.26   & \textbf{25.93}    &\textbf{19.74} & 14.42 & \textbf{31.96} & 26.67 \\ 
 & LoRA   & \textbf{37.09}     & 22.13   &\textbf{67.31}  &59.02 &\textbf{18.08} & 15.81 &   \textbf{23.54}   &     19.85           & 24.93       & \textbf{25.65}    &\textbf{19.27} & 13.79 & \textbf{31.70} & 26.04 \\ 
\midrule
\multirow{4}{*}{\textbf{Small}} & Zero-shot        & 11.65      & \textbf{18.77}      &29.88 &\textbf{38.42} & 7.07 &   \textbf{7.12}    &    12.31          & \textbf{12.42}       & 0.00  &\textbf{1.01} & \textbf{12.35}& 9.25 & 12.21 &\textbf{14.50} \\ 
& P-tuning     & \textbf{39.27}       & 18.58   &\textbf{70.31}   &61.13 &\textbf{19.98} & 15.13 &    \textbf{24.75}   &     20.34          & 22.77       & \textbf{24.11}    &\textbf{18.76} &14.61 &\textbf{32.64} & 25.65  \\
 &Adapter         & \textbf{42.29}       & 23.68   &\textbf{73.21}   &64.91 &\textbf{21.40} & 18.10 &  \textbf{27.29}  &     20.55         & 24.93    &\textbf{25.07}    &20.17 &15.03 &\textbf{34.88} & 27.89 \\ 
 & LoRA   & \textbf{39.27}     & 28.06   &\textbf{72.34} &64.59 &\textbf{20.98} & 17.90 &   \textbf{24.83}   & 21.28             & \textbf{25.79}      & 24.69   &\textbf{20.31}& 14.46 & \textbf{33.92} & 28.50 \\ 
  & Fine-tuning  &  \textbf{36.27}    & 21.87   & \textbf{73.83}  &63.05 &\textbf{17.80} & 13.11 &  \textbf{30.84}    &     21.26          &      \textbf{26.08}  & 25.79    & \textbf{20.79} & 14.79 &\textbf{34.27}  & 26.65 \\
\midrule
\multirow{5}{*}{\textbf{Medium}} & Zero-shot        & 23.67       & \textbf{24.11}   & 51.00   &\textbf{52.17} &\textbf{8.90} &6.39 &   9.01    &     \textbf{10.04}           & \textbf{1.44}       & 0.14    &\textbf{11.28}& 9.15& \textbf{17.55} &17.00 \\ 
& P-tuning      & \textbf{45.52}       &24.18    &\textbf{77.00}   &67.94 &\textbf{24.50} & 19.02 &    \textbf{33.31}    &     24.20          & \textbf{32.74}      & 31.93    & \textbf{20.37} & 15.40 & \textbf{38.91} & 30.44  \\
 &Adapter      & \textbf{46.71}     &43.01   &\textbf{78.05}  &71.35 &\textbf{24.30} & 20.51 &    \textbf{32.53}    &     25.90        & \textbf{40.84}   & 31.98    &\textbf{20.03} & 15.61 & \textbf{40.41} & 34.65 \\ 
 & LoRA  & \textbf{46.81}     & 42.11   & \textbf{78.26}  &70.75 &\textbf{25.17} & 20.42 &   \textbf{31.84}   &     24.48           & \textbf{41.56}       & 32.41    &\textbf{21.47} & 15.30 & \textbf{40.85} & 34.24 \\ 
 & Fine-tuning  &  \textbf{41.34}    & 29.79   & \textbf{79.82}  &68.84 &\textbf{22.33} & 19.37 &  \textbf{49.67}    &     23.53          &      \textbf{37.01}  & 33.56    & \textbf{21.95} & 15.29 &\textbf{42.02}  & 31.73 \\
\midrule
\multirow{4}{*}{\textbf{Large}} & Zero-shot        & 25.37       & \textbf{31.43}   & 48.68   &\textbf{60.30} &\textbf{13.92} &7.98 &   8.73   &     \textbf{10.52}           & \textbf{2.97}       & 1.87    & 6.30& \textbf{9.33}& 17.66 &\textbf{20.24} \\ 
& P-tuning      & \textbf{45.20}       & 15.78   & \textbf{78.33}   &73.22 & \textbf{25.21} & 21.58 &    \textbf{34.24}    &     24.50          & \textbf{47.65}      & 39.93    & \textbf{20.07} & 15.00 &\textbf{41.78} & 31.67  \\
 &Adapter        & \textbf{47.48}     &44.43   & \textbf{79.68}  &73.57 & \textbf{26.37} & 22.03 &    \textbf{32.12}    &     26.09        & \textbf{46.74}   & 38.06    & \textbf{20.81} & 15.22 & \textbf{42.20} & 36.57 \\ 
 & LoRA  & \textbf{47.33}     & 44.48  & \textbf{79.79}  &73.63 & \textbf{25.85} & 21.49 &   \textbf{32.25}   &     25.21           & \textbf{42.62}       & 39.31    & \textbf{21.67} & 15.02 & \textbf{41.58} & 36.53 \\ 
 \midrule
\multirow{4}{*}{\textbf{Extra Large}} & Zero-shot        & 26.97       & \textbf{33.49}   & 44.71   &\textbf{62.87} & \textbf{11.89} &10.07 &   11.58    &     \textbf{13.38}           & \textbf{3.07}       & 0.96    & 7.65& \textbf{9.99}& 17.65 &\textbf{21.79} \\ 
& P-tuning  & \textbf{47.27}       &24.53    & \textbf{80.27}   &74.38 & \textbf{27.09} & 22.48 &    \textbf{34.08}    &     24.93      & \textbf{57.19}      & 38.06    &\textbf{21.17} & 15.53 & \textbf{44.51} & 33.32  \\
 &Adapter         & \textbf{49.68}     &46.41   & \textbf{81.64}  &75.10 & \textbf{26.94} & 22.24 &    \textbf{33.94}    &     26.38        & \textbf{54.65}   & 42.62    & \textbf{21.19} & 15.71 & \textbf{46.82} & 38.08 \\ 
 & LoRA  & \textbf{49.21}     & 44.53   & \textbf{81.87}  &74.92 & \textbf{27.31}& 22.16 &   \textbf{31.98}   &     27.49           & \textbf{49.19}       & 39.65    & \textbf{22.77} & 15.73 & \textbf{43.72} & 37.41 \\ 
\bottomrule
\end{tabular}
}
\caption{A comprehensive comparison between GPT vs RETRO on six datasets. \textbf{Bold} indicates the better result in each head-to-head comparison.}
\label{tab:main}
\vspace{-4mm}
\end{table*}

\section{Results}

\subsection{Main Results}

Table~\ref{tab:main} shows the comprehensive comparison between GPT and RETRO models across five model sizes and six datasets. We perform zero-shot and PEFT on all cases and fine-tuning on small and medium model sizes. From this table we observe:

\textbf{1) RETRO is better than GPT at zero-shot retrieval tasks.} This superiority stems from its unique pre-training approach and focus on retrieval tasks. By learning to extract salient information from retrieved text and integrate it into its generation process, RETRO develops the capability to harness relevant contextual knowledge, ultimately leading to its strong zero-shot performance. In contrast, GPT relies on an auto-regressive loss during pre-training, focusing on accurately predicting next tokens without the benefit of external retrievals. As a result, GPT's ability to learn context-aware question-answering is limited to the presence of relevant data within the pre-training corpus, resulting in less targeted training compared to RETRO.

\textbf{2) Both RETRO and GPT models exhibit saturation points around 8B parameters}. Additionally, a similar pattern emerges between the two models as they are scaled, albeit with RETRO performing less well. This can be seen in Figure \ref{fig:peft_methods} and suggests that, for a specific task, a medium-sized PEFT model strikes the optimal balance between cost and performance, making it a sweet spot for many applications.

\begin{figure}[ht]
    \centering
    \hspace{-0.49cm}
    \includegraphics[scale=0.54]{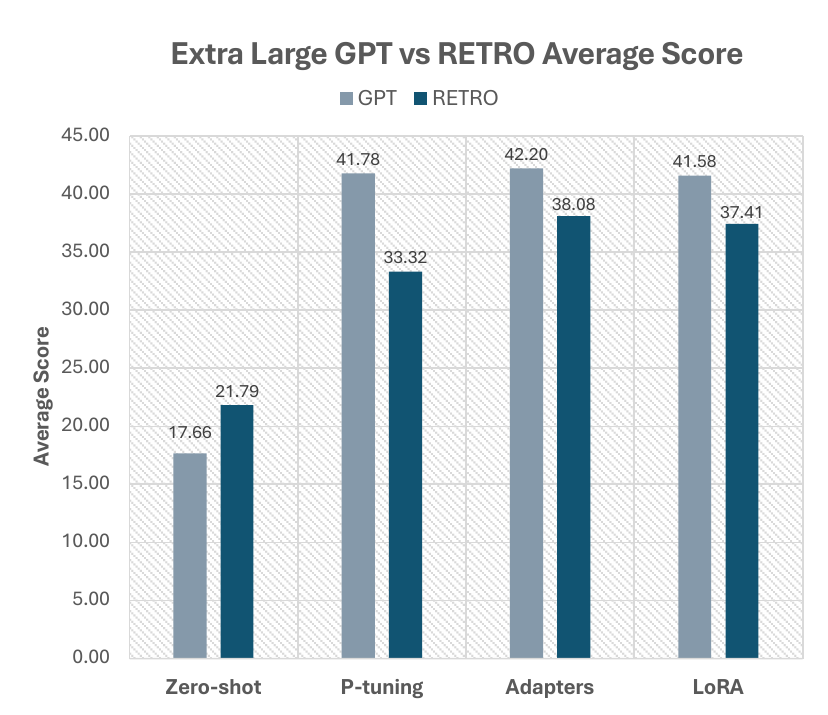}
    \caption{Comparison of Extra Large GPT and RETRO results averaged across 6 datasets.}
    \label{fig:43B_comparison}
    \vspace{-4mm}
\end{figure}

\textbf{3) P-tuning underperforms LoRA and Adapters in GPT and RETRO models for these tasks.} This difference is visualized in Figure \ref{fig:43B_comparison} and Figure \ref{fig:small_medium_compare} (Appendix~\ref{sec:appendixc}). For RETRO models, P-tuning under performs the other PEFT methods more significantly. We believe that P-Tuning's lower parameter count contributes to its lower performance especially when paired with smaller base model sizes. For RETRO P-Tuning specifically, we hypothesis that P-Tuning's weaker ability in all RETRO model sizes could lie in architecture differences. In P-tuning, virtual tokens are intentionally prepended to the decoder’s input, but they are not included in the retrieval encoder. Although they can influence the encoder through cross-attention, the impact might not be as direct or substantial as required. Alternatively, LoRA and Adapters are added to both encoder and decoder which explains their improved capabilities.

\textbf{4) The performance ceiling for PEFT-tuned models is notably higher for GPT than RETRO}. This is demonstrated in Figure \ref{fig:individual_comparison} (Appendix~\ref{sec:appendixc}) where for example, with medium-sized models, the average score of LoRA with GPT is 40.85, while with RETRO it is 34.24. This disparity is seen in all the PEFT methods across multiple model sizes and suggests that GPT is better suited for PEFT. This phenomenon can also be possibly explained by the two different pre-training strategies. Since GPT pre-training is not focused on retrieval-augmented generation, it allows for a larger room for improvement when fine-tuned on these tasks.

\textbf{5) Full fine-tuning marginally outperforms PEFT in GPT models and underperforms in RETRO models.} We find that full fine-tuning in GPT models achieves slightly better performance than PEFT on 4 out of 6 tasks while RETRO slightly underperforms on 5 out of 6 tasks. Interestingly, NQ and NQA underperforms against PEFT in both GPT and RETRO 2B and 8B model sizes while both model sizes see notable improvements in fine-tuning GPT on the QASPER dataset. This aligns with previous findings \citep{lora}, potentially because PEFT serves as a regularization, forcing models to learn better.

\vspace{-3.5mm}

\subsection{Failure Case Analysis}

To better frame and qualitatively understand our results we study on an entry from the NQ test set evaluated with Extra-Small RETRO model. Figure \ref{fig:sample_entry} demonstrates how zero-shot RETRO is capable of achieving the correct answer but incorrectly formatting the output. Contrarily, P-Tuning incorrectly hallucinates an answer of "1947", the first date seen in the context. LoRA achieves the desired answer by correctly parsing the context and formatting with the desired brevity.
\subsection{Comparing to Instruction-tuned RETRO}

Instruction tuning post retrieval-augmented pre-training~\citep{instructretro} has been demonstrated to improve zero-shot performance on RETRO models. A natural thought is that whether Instruction-tuned RETRO (I-RETRO) serve as a better foundation for applying PEFT compared to the base RETRO. To investigate this, we additionally apply PEFT to a medium-sized I-RETRO model and show overall results in Table~\ref{tab:i-retro-short} and more granular results in Table~\ref{tab:i-retro-long} (Appendix~\ref{sec:appendixc}). Our findings reveal that while I-RETRO exhibits improved performance in the zero-shot setting, it has limited scope for further improvement using PEFT. Even with substantial hyperparameter tuning, the average scores across six datasets, using each of the three PEFT methods, demonstrate an approximately 10\% gap between I-RETRO and base RETRO. We hypothesize that conceptually both models should be tunable to similar performance but will leave that exploration to future work.

\begin{table}[ht]
\centering
\renewcommand{\arraystretch}{1.5}
\Huge
\resizebox{\columnwidth}{!}{
\begin{tabular}{lccccccccccc}
\toprule
                             & \multicolumn{2}{c}{\textbf{Average QA}} &
                            \multicolumn{2}{c}{\textbf{QUALITY}} &
                            \multicolumn{2}{c}{\textbf{QMSUM}}&
                            \multicolumn{2}{c}{\textbf{Average}} \\                           
                             & I-RETRO & RETRO & I-RETRO & RETRO & I-RETRO & RETRO & I-RETRO & RETRO         \\
\midrule
 Zero-shot        &  \textbf{27.65}      &  23.79  & \textbf{3.35}    &0.14 & \textbf{11.04} & 9.15 & \textbf{20.83}  & 17.00 \\ 
 P-tuning     &  23.25      & \textbf{47.18}           &  16.68   & \textbf{31.93}    & \textbf{15.88} & 15.40 &20.75  & \textbf{30.44}  \\
 Adapter         &22.64      &\textbf{52.75}        &  29.87  & \textbf{31.98}    & 15.06 & \textbf{15.16} & 22.58 & \textbf{34.65} \\ 
  LoRA   &  26.53   & \textbf{52.80}     &   24.21     & \textbf{32.41}    & \textbf{15.40} & 15.30 & 24.29 & \textbf{34.24} \\ 

\bottomrule
\end{tabular}
}
\caption{Instruction-tuned RETRO evaluation results.}
\label{tab:i-retro-short}
\vspace{-4mm}
\end{table}

\section{Conclusion}

This study explores Parameter-Efficient Fine-Tuning (PEFT) methods applied to Retrieval-Augmented Generation (RAG) models, comparing GPT and RETRO architectures. RETRO generally outperforms GPT in zero-shot settings due to their pre-training process that integrates external retrieval, enhancing contextual understanding. However, GPT models show a higher performance potential with PEFT, indicating more room for improvement during fine-tuning. Both RETRO and GPT models perform optimally around the 8B parameter mark, balancing cost and performance. While P-tuning is effective in larger models, it lags behind other methods in smaller models, particularly for RETRO. Applying PEFT to Instruction-tuned RETRO yields limited improvement compared to base RETRO, suggesting a saturation point in leveraging pre-training and fine-tuning benefits. Our comprehensive analysis offers valuable insights for optimizing large language models with PEFT and RAG to the community.

\section*{Limitations}

Due to the breadth of experiments covered in this work we had to prioritze certain experiments over others. This resulted in us using only the small and medium sized GPT and RETRO models for additional finetuning and Instruction tuning experiments. We believe these results generalize to the other model sizes but leave that to be validated in future work.

\section*{Potential Risks}

The environmental impact associated with training and fine-tuning large models is not negligible as it involves substantial computational resources and energy consumption. While PEFT aims to alleviate this by reducing the number of tunable parameters, works like ours still require significant compute to distinguish which methods are more promising.

\bibliography{latex/custom}

\appendix

\section{Details on Experimental Setup}
\label{sec:appendixa}

\subsection{Hyperparameter Tuning}
\label{sec:hyperparam_tuning}

Given the massive number of experiments required for this work, we used an initial search of learning rates 1e-4 and 1e-5 followed by selectively modifying certain hyperparameters if a model, method and dataset combination did not converge. For all experiments we used a micro batch size of 1 and global batch size of 32 or 128 using tensor parallelism combined with a max sequence length of 1024 and 5 retrieved neighbors. For P-Tuning we selected 100 virtual tokens, kept dropout at 0.0 and used 2 multilayer perceptron layers with hidden sizes of 2048 as the prompt encoder. For Adapters/LoRA we used 32 and 64 dimensions with parallel type adapters and kept dropout at 0.0. In certain runs on NQ and TQA datasets we noticed the models did not converge. To address this, we conducted additional hyperparameter search by varying the learning rates between 1e-4 and 1e-6, testing P-Tuning with 40, 50, and 90 virtual tokens, and selecting Adapters/LoRA with a dimension of 16.

\subsection{Resource Utilization}
\label{sec:resource_util}

In our experiments, we used up to 16 compute nodes, each with 8 A100-80GB SXM GPUs. When model is smaller, we increased the data parallelism size, using tools in \href{https://github.com/NVIDIA/NeMo}{NeMo framework}. 

\subsection{Prompt Template}
\label{sec:prompt-template}

The template we used to present context to GPT models is as follows.

\begin{verbatim}
title: {title}
source: {source}
title: {title}
source: {source}
title: {title}
source: {source}
title: {title}
source: {source}
title: {title}
source: {source}

Question: {question} Answer: The answer is
\end{verbatim}

\section{Supplementary Figures and Tables}
\label{sec:appendixc}
\vspace{-5mm}

\begin{figure*}[ht]
        \subfloat{
            \includegraphics[width=.5\linewidth]{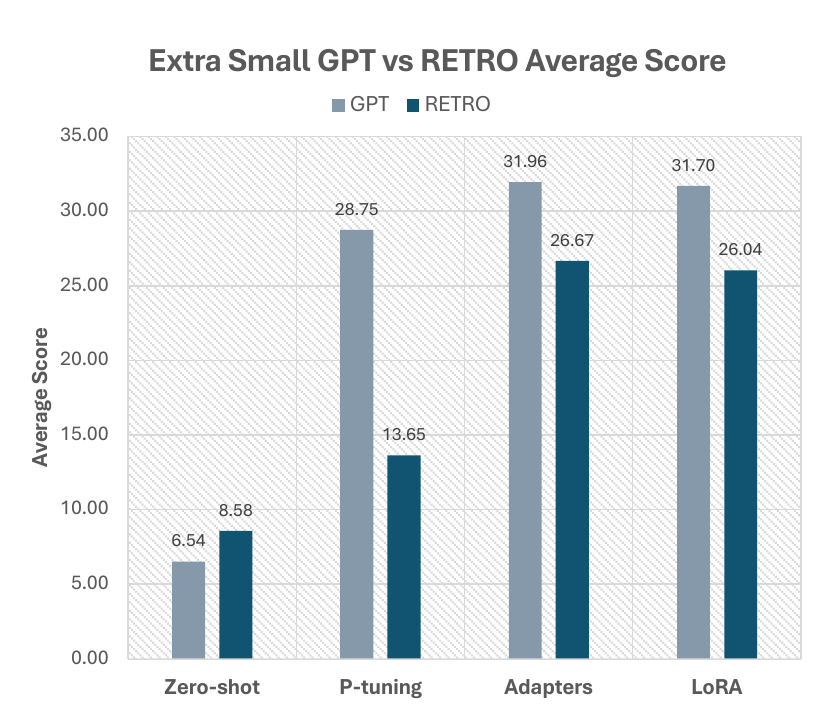}
            \label{subfig:a}
        }\hfill
        \subfloat{
            \includegraphics[width=.5\linewidth]{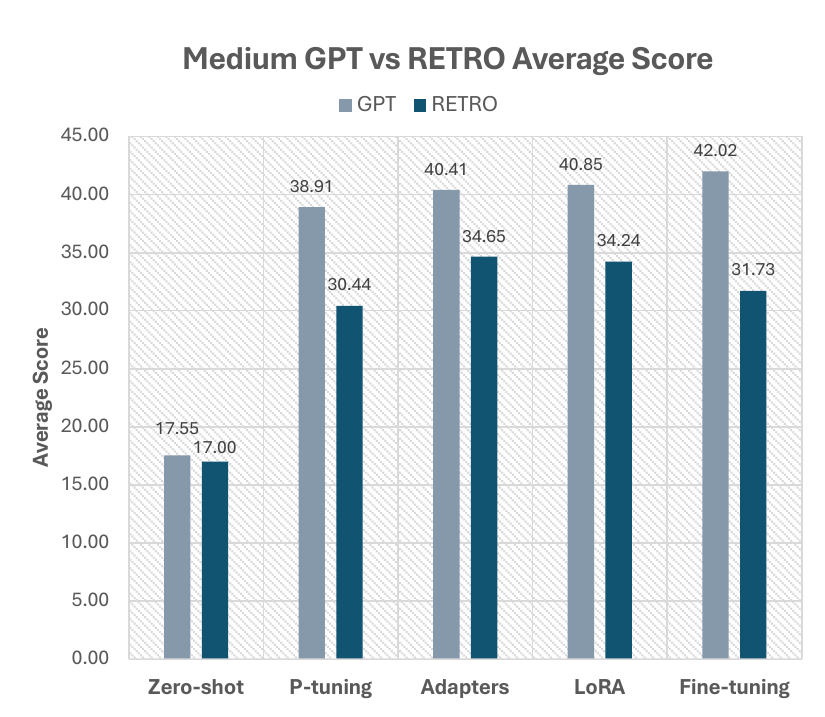}
            \label{subfig:b}
        }\\
            \vspace{-3.5mm}
        \caption{GPT vs RETRO comparisons on Extra Small and Medium sized models.}
        \label{fig:small_medium_compare}
            \vspace{-3.5mm}
\end{figure*}

\begin{figure*}[!h]
        \subfloat{
            \includegraphics[width=.5\linewidth]{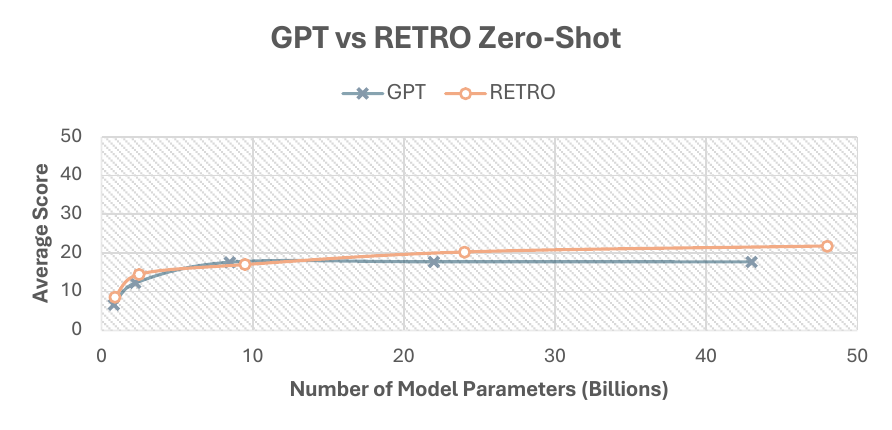}
            \label{subfig:a}
        }\hfill
        \subfloat{
            \includegraphics[width=.5\linewidth]{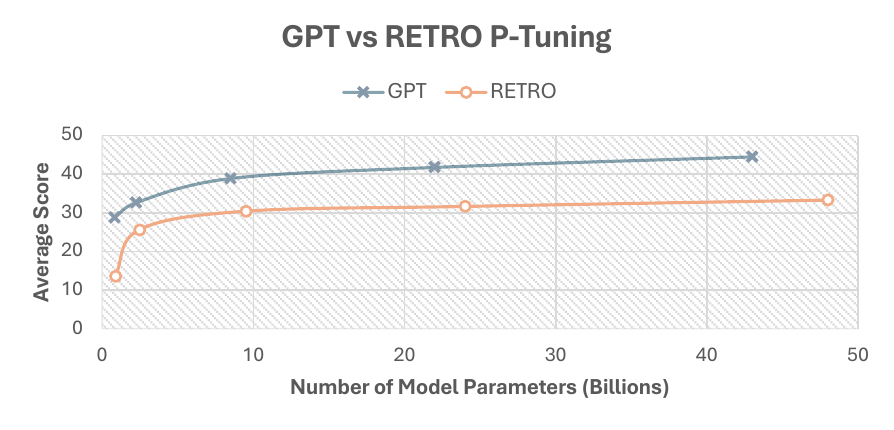}
            \label{subfig:b}
        }\\
        \subfloat{
            \includegraphics[width=.5\linewidth]{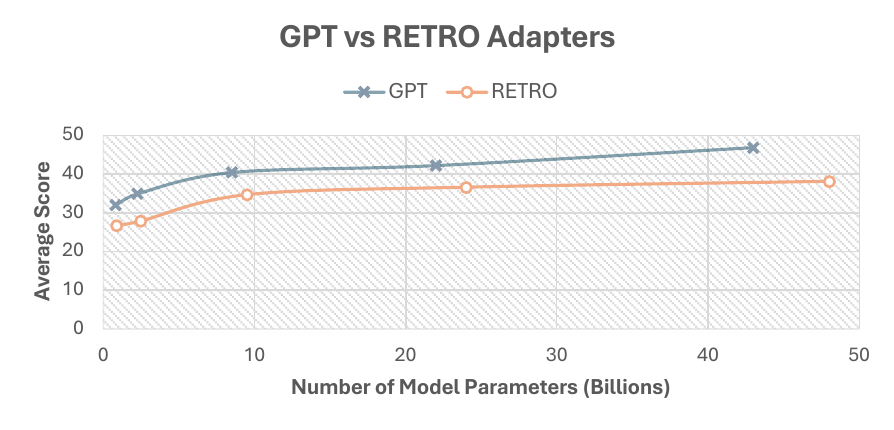}
            \label{subfig:c}
        }\hfill
        \subfloat{
            \includegraphics[width=.5\linewidth]{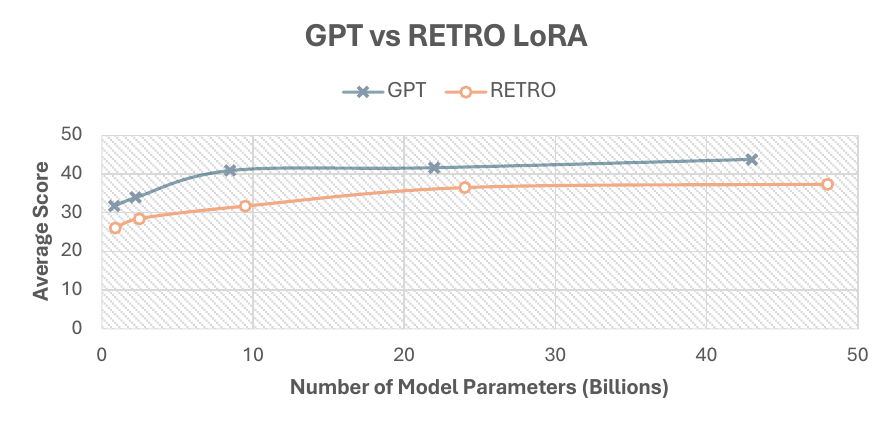}
            \label{subfig:d}
        }
            \vspace{-3.5mm}
        \caption{GPT vs RETRO seperate method comparisons.}
        \label{fig:individual_comparison}
    \vspace{-3.5mm}
\end{figure*}

\begin{table*}[h]
\centering
\renewcommand{\arraystretch}{1.5} 
\Huge 
\resizebox{\textwidth}{!}{
\begin{tabular}{lccccccccccccccccc}
\toprule
                             & \multicolumn{2}{c}{\textbf{NQ}} &
                            \multicolumn{2}{c}{\textbf{TQA}}&
                            \multicolumn{2}{c}{\textbf{NQA}} &
                            \multicolumn{2}{c}{\textbf{QASPER}} & 
                            \multicolumn{2}{c}{\textbf{QUALITY}} &
                            \multicolumn{2}{c}{\textbf{QMSUM}}&
                            \multicolumn{2}{c}{\textbf{Average}} \\                           
                             & I-RETRO & RETRO & I-RETRO & RETRO & I-RETRO & RETRO & I-RETRO & RETRO & I-RETRO & RETRO & I-RETRO & RETRO & I-RETRO & RETRO         \\
\midrule
 Zero-shot        &  \textbf{30.39}      &  24.11  &  \textbf{53.25}  &52.17 &\textbf{12.23} &  6.39    &  \textbf{14.72}           &    10.04   & \textbf{3.35}    &0.14 & \textbf{11.04} & 9.15 & \textbf{20.83}  & 17.00 \\ 
 P-tuning     &  19.55      & \textbf{24.18}   & 41.95  & \textbf{67.94} & \textbf{20.17} & 19.02 &   11.34    &     \textbf{24.20}          &  16.68   & \textbf{31.93}    & \textbf{15.88} & 15.40 &20.75  & \textbf{30.44}  \\
 Adapter         &18.81      &\textbf{43.01}   &38.83  &\textbf{71.35} &  20.30 & \textbf{20.51} &  12.64  &     \textbf{25.90}        &  29.87  & \textbf{31.98}    & 15.06 & \textbf{15.16} & 22.58 & \textbf{34.65} \\ 
  LoRA   &  21.56   & \textbf{42.11}   & 47.89 &\textbf{70.75} & 19.23 & \textbf{20.42} &  17.45    &     \textbf{24.48}           &   24.21     & \textbf{32.41}    & \textbf{15.40} & 15.30 & 24.29 & \textbf{34.24} \\ 

\bottomrule
\vspace{-3.5mm}
\end{tabular}
}
    \vspace{-3.5mm}
\caption{Full results with Instruction-tuned RETRO. \textbf{Bold} indicates the better result in each head-to-head comparison.}
\label{tab:i-retro-long}
\vspace{-3.5mm}
\end{table*}

\end{document}